\documentclass[letterpaper]{article}
\usepackage{aaai2026}
\usepackage{mathptmx}
\usepackage{newtxmath} 
\usepackage{helvet}
\usepackage{courier}
\usepackage[hyphens]{url}
\usepackage{graphicx}
\usepackage{natbib}
\usepackage{caption}
\usepackage{array}  
\usepackage{multirow} 
\usepackage{graphicx} 
\usepackage{amsmath}

\usepackage{algorithm}
\usepackage{algorithmic}
\usepackage{newfloat}
\usepackage{listings}

\DeclareCaptionStyle{ruled}{labelfont=normalfont,labelsep=colon,strut=off}
\floatstyle{ruled}
\newfloat{listing}{tb}{lst}{}
\floatname{listing}{Listing}

\title{TTF-VLA: Temporal Token Fusion via Pixel-Attention Integration for Vision-Language-Action Models}

\author{
    Chenghao Liu\textsuperscript{\rm 1}\thanks{Equal contribution.},
    Jiachen Zhang\textsuperscript{\rm 1}\footnotemark[1],
    Chengxuan Li\textsuperscript{\rm 1}\footnotemark[1],
    Zhimu Zhou\textsuperscript{\rm 1},
    Shixin Wu\textsuperscript{\rm 1},\\
    Songfang Huang\textsuperscript{\rm 1}\thanks{Corresponding authors.},
    Huiling Duan\textsuperscript{\rm 1}\footnotemark[2]
}

\affiliations{
    \textsuperscript{\rm 1}School of Advanced Manufacturing and Robotics, Peking University\\
    chliu@stu.pku.edu.cn, z89498323286@gmail.com, lichengxuan0904@gmail.com, z18082717992@gmail.com, wushixin@stu.pku.edu.cn, hsf@pku.edu.cn, hlduan@pku.edu.cn
}

\begin{document}

\maketitle

\begin{abstract}
Vision-Language-Action (VLA) models process visual inputs independently at each timestep, discarding valuable temporal information inherent in robotic manipulation tasks. This frame-by-frame processing makes models vulnerable to visual noise while ignoring the substantial coherence between consecutive frames in manipulation sequences. We propose Temporal Token Fusion (TTF), a training-free approach that intelligently integrates historical and current visual representations to enhance VLA inference quality. Our method employs dual-dimension detection combining efficient grayscale pixel difference analysis with attention-based semantic relevance assessment, enabling selective temporal token fusion through hard fusion strategies and keyframe anchoring to prevent error accumulation. Comprehensive experiments across LIBERO, SimplerEnv, and real robot tasks demonstrate consistent improvements: 4.0 percentage points average on LIBERO (72.4\% vs 68.4\% baseline), cross-environment validation on SimplerEnv (4.8\% relative improvement), and 8.7\% relative improvement on real robot tasks. Our approach proves model-agnostic, working across OpenVLA and VLA-Cache architectures. Notably, TTF reveals that selective Query matrix reuse in attention mechanisms enhances rather than compromises performance, suggesting promising directions for direct KQV matrix reuse strategies that achieve computational acceleration while improving task success rates.
\end{abstract}

\begin{links}
    \link{Code}{https://github.com/PKU-XLab/TTF-VLA}
\end{links}

\section{Introduction}

Vision-Language-Action (VLA) models have emerged as a transformative paradigm in robotic manipulation, seamlessly integrating visual perception, natural language understanding, and action generation within unified neural architectures. Building upon the success of large-scale vision-language transformers, recent VLA systems~\cite{brohan2022rt1,brohan2023rt2,pertsch2024octo,kim2024openvla,kim2025oft,black2024pi0,black2025pi05} have demonstrated unprecedented capabilities in executing complex manipulation instructions across diverse environments. These models fundamentally reshape robotic control by treating action prediction as a multimodal sequence generation task, where visual observations and natural language instructions are jointly processed to produce discrete action tokens that guide continuous robot behavior.

However, despite their remarkable achievements, current VLA models suffer from a critical limitation: they process visual inputs in temporal isolation, treating each frame independently without leveraging the substantial temporal coherence inherent in robotic manipulation sequences. This frame-by-frame processing systematically recomputes all visual tokens from scratch at each timestep, discarding valuable temporal information even when the majority of visual content remains consistent across adjacent frames. Moreover, this approach makes models vulnerable to visual noise including lighting fluctuations, motion blur, and sensor artifacts that are common in robotic manipulation environments.

This temporal myopia creates a fundamental challenge: while naive historical token integration risks overlooking critical changes in object poses or environmental conditions, completely ignoring temporal context misses opportunities to leverage the structured patterns inherent in robotic manipulation. Specifically, visual changes typically concentrate in localized, task-relevant regions while background areas remain static. This observation suggests that effective temporal integration requires distinguishing between \textit{spatial dynamics} from physical movements and \textit{semantic relevance} shifts reflecting task-specific importance.

Motivated by these insights, we propose a training-free temporal token fusion framework that intelligently integrates historical and current visual representations to enhance VLA inference quality. Our approach introduces a dual-dimension detection mechanism that combines computationally efficient grayscale pixel difference analysis with attention-guided semantic relevance assessment, enabling informed decisions about temporal token integration. Through hard fusion strategies coupled with adaptive keyframe mechanisms, our method effectively balances temporal coherence with responsiveness to task-critical changes while preventing long-term error accumulation.

Our key contributions are:
\begin{itemize}
\item A novel temporal token fusion framework featuring dual-dimension detection that combines efficient grayscale pixel difference analysis with attention-guided semantic relevance assessment for intelligent integration of historical and current visual representations.
\item An adaptive fusion strategy employing hard token selection with keyframe anchoring mechanisms that balances temporal coherence with responsiveness to task-critical changes, applicable across diverse VLA architectures without requiring model retraining.
\item Comprehensive experimental validation: 4.0 percentage points average improvement on LIBERO~\cite{liu2023libero}, cross-environment generalization on SimplerEnv~\cite{li24simpler}, and 8.7\% relative improvement on real robot tasks, with model-agnostic applicability across OpenVLA~\cite{kim2024openvla} and VLA-Cache~\cite{xu2025vla} architectures.
\item Discovery of beneficial Query matrix reuse through temporal token fusion, revealing promising ``free lunch'' directions for direct KQV matrix reuse that achieve computational acceleration while improving task performance.
\end{itemize}

Our approach provides a principled solution to the fundamental tension between leveraging temporal coherence and maintaining sensitivity to dynamic changes in VLA models.

\section{Related Work}

\noindent\textbf{Vision-Language-Action Models} Vision-Language-Action (VLA) models have emerged as a unified framework for robotic manipulation, integrating visual perception, language understanding, and action prediction. Early works like RT-1 and RT-2 tokenized actions with visual and linguistic inputs for end-to-end policy learning~\cite{brohan2022rt1,brohan2023rt2}. Recent advancements include Octo and OpenVLA for open-source implementations~\cite{pertsch2024octo,kim2024openvla,kim2025oft}, and Pi-0/Pi-0.5 for flow-based architectures~\cite{black2024pi0,black2025pi05}. However, current models exhibit limitations in handling temporal dynamics and visual artifacts~\cite{chi2023diffusion,wen2024diffusion,openx2023,huang2024a3vlm}, often processing frames independently while ignoring temporal coherence.

\noindent\textbf{Token Processing Techniques} Efficient token processing has become crucial for transformer deployment. In vision transformers, attention-guided methods such as DynamicViT, AdaViT, and EViT progressively prune redundant tokens~\cite{dynamicvit2021,adavit2022,evit2022}, while ToMe optimizes efficiency through strategic token consolidation~\cite{tome2023}. For vision-language models, recent approaches like FastV and SparseVLM focus on visual token sparsification using text-guided selection strategies~\cite{chen2024image,zhang2024sparsevlm}. VLA-Cache specifically targets robotic scenarios through KV-cache reuse mechanisms~\cite{xu2025vla}. However, these methods primarily address spatial redundancy within individual frames. Unlike these spatial compression approaches, our work targets temporal redundancy across sequential frames, proposing a dual-dimension strategy that leverages historical information to enhance VLA inference quality.

\section{Methodology}

\begin{figure*}[t]
\centering
\includegraphics[width=0.9\textwidth]{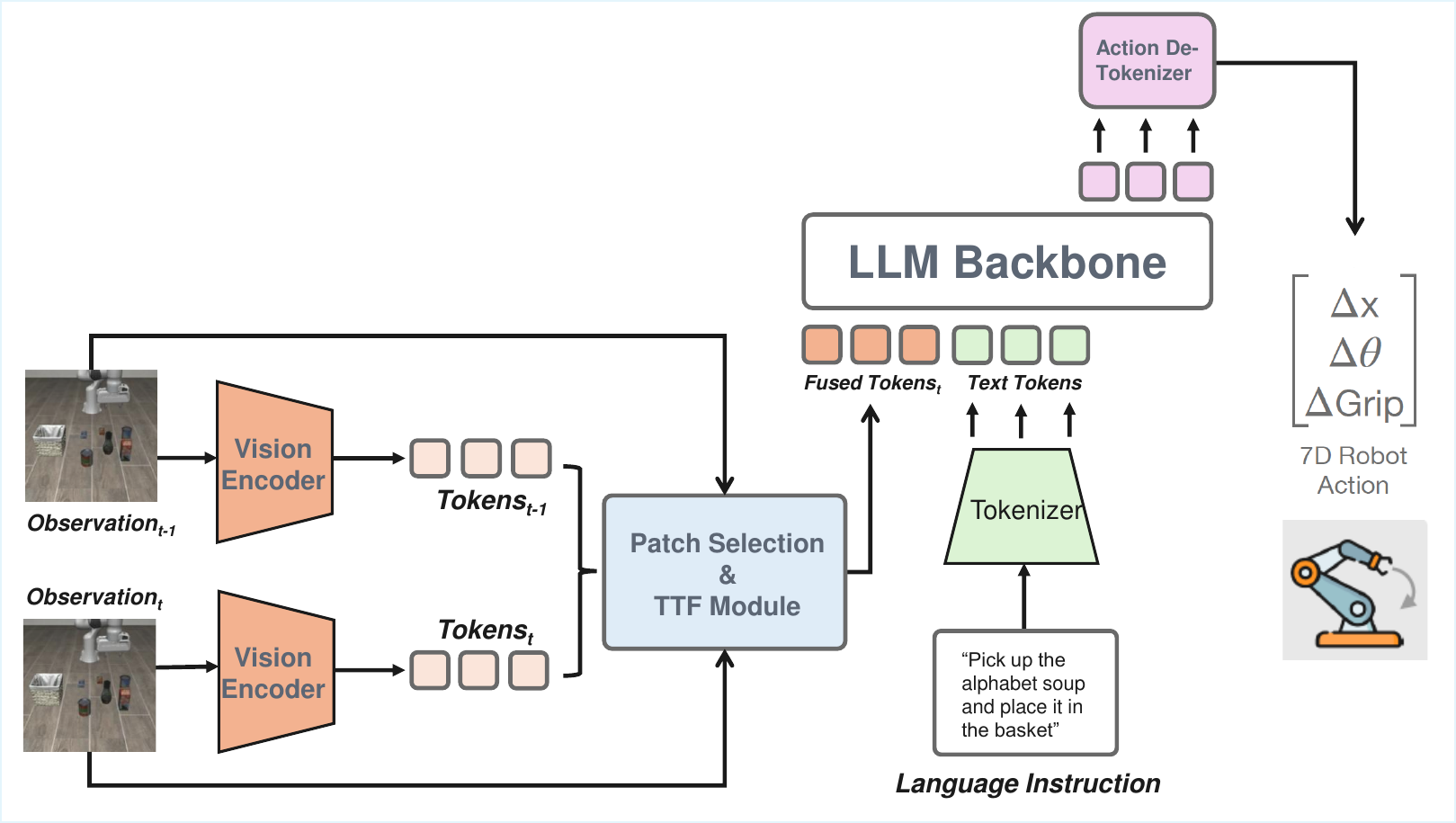}
\caption{\textbf{Overall Framework of Temporal Token Fusion for VLA Models.} The framework illustrates the end-to-end process, where the Vision Encoder extracts tokens from current (Observation$_t$) and previous (Observation$_{t-1}$) frames. These are then processed by the Patch Selection module and TTF module for patch selection and token fusion. The fused tokens are subsequently fed into the LLM Backbone, combined with language instruction, to generate 7-DoF robotic actions via the Action Detokenizer.}
\label{fig:overall_framework}
\end{figure*}

In robotic manipulation tasks, consecutive frames often exhibit substantial visual redundancy, yet subtle but critical changes in object poses, lighting conditions, or environmental context can significantly impact action prediction quality. Our approach leverages temporal token fusion to enhance VLA inference quality through intelligent integration of historical and current visual representations. Figure~\ref{fig:overall_framework} provides an overview of our complete framework.

\subsection{Problem Formulation}

Vision-Language-Action models process sequential inputs of the form $\{\mathbf{I}_t, \mathbf{L}_t\} \rightarrow \mathbf{A}_t$, where $\mathbf{I}_t \in \mathbb{R}^{H \times W \times C}$ represents the visual observation, $\mathbf{L}_t$ denotes the language task instruction, and $\mathbf{A}_t \in \mathbb{R}^7$ represents the predicted 7-DoF robotic action at timestep $t$. 

Current VLA models typically employ vision-language transformer architectures that process visual observations through patch-based encoders. The vision encoder extracts patch tokens $\mathbf{T}_t = \{\mathbf{t}_t^{(i)}\}_{i=1}^{N}$ from input images, which are then projected to the language model's embedding space and integrated with tokenized task instructions before being processed by the transformer backbone.

\subsubsection{Temporal Token Fusion} Given a sequence of visual observations $\{\mathbf{I}_1, \mathbf{I}_2, \ldots, \mathbf{I}_t\}$ and corresponding patch tokens $\{\mathbf{T}_1, \mathbf{T}_2, \ldots, \mathbf{T}_t\}$, our goal is to learn a fusion function $\mathcal{F}$ that intelligently integrates temporal information:

\begin{equation}
\tilde{\mathbf{T}}_t = \mathcal{F}(\mathbf{T}_t, \mathbf{T}_{t-1}, \mathbf{I}_t, \mathbf{I}_{t-1}, \mathbf{L}_t)
\end{equation}

where $\tilde{\mathbf{T}}_t$ represents the temporally-fused patch tokens that maintain critical current information while leveraging relevant historical context. The fusion function must balance temporal coherence with responsiveness to important changes in the visual scene. 

\subsection{Temporal Token Fusion Framework}

\subsubsection{Hard Fusion Strategy}
Our temporal fusion framework operates through a systematic process that evaluates each patch for intelligent temporal integration, as detailed in Figure~\ref{fig:patch_selection}. The framework employs a hard fusion strategy that makes binary selection decisions for each patch, choosing between current and historical tokens based on dual-dimension detection:

\begin{equation}
\tilde{\mathbf{t}}_t^{(i)} = \begin{cases}
\mathbf{t}_t^{(i)} & \text{if } m_i^{\text{fusion}} = 1 \\
\mathbf{t}_{t-1}^{(i)} & \text{if } m_i^{\text{fusion}} = 0
\end{cases}
\end{equation}

where $m_i^{\text{fusion}} \in \{0,1\}$ is the binary fusion mask that determines temporal integration decisions for patch $i$, computed through our dual-dimension detection combining grayscale pixel difference detection and attention-based semantic relevance detection (detailed in Section 3.3). This strategy provides clear temporal context selection, aligning with the discrete nature of robotic manipulation tasks, where important patches (mask=1) use current frame tokens and others (mask=0) reuse previous frame tokens.

\subsubsection{Keyframe Mechanism}
To prevent long-term error accumulation, we introduce periodic keyframes where all patches are unconditionally recomputed:
\begin{equation}
\text{IsKeyframe}(t) = (t \bmod K = 0) \lor (\mathbf{T}_{t-1} = \emptyset)
\end{equation}

The keyframe interval $K$ balances temporal context with stability, preventing long-term error accumulation while preserving temporal coherence benefits.

The final fusion decision integrates both dimensions through carefully designed combination rules. The fusion mask $m_i^{\text{fusion}} \in \{0,1\}$ determines the final temporal integration decision for patch $i$, where 1 indicates using current tokens for important patches and 0 indicates reusing historical tokens for others, as guided by the dual-dimension detection in Figure~\ref{fig:patch_selection}(a). We use logical OR to ensure patches are updated when exhibiting either type of change:
\begin{equation}
m_i^{\text{fusion}} = m_i^{\text{pixel}} \lor m_i^{\text{attention}}
\end{equation}

This OR operation ensures that patches use current frame tokens when either dimension indicates importance, providing conservative fusion that prioritizes inference quality through comprehensive temporal context integration.

\subsection{Dual-Dimension Detection}

\begin{figure}[t]
\centering
\includegraphics[width=\columnwidth]{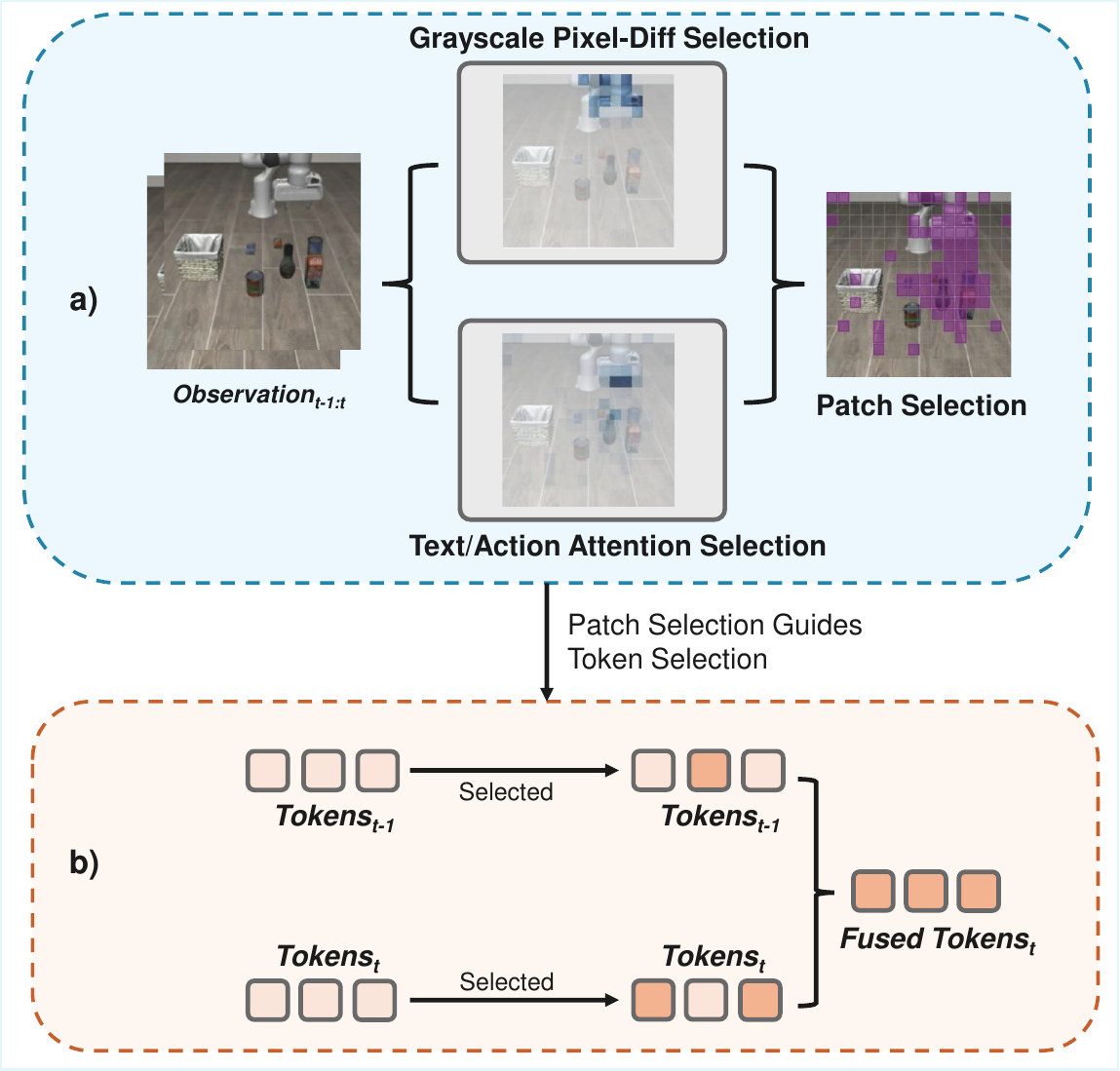}
\caption{\textbf{The Details of Patch Selection and Temporal Token Fusion.} The process includes (a) Grayscale Pixel Difference Detection and Attention-Based Semantic Relevance Detection for identifying important patches, and (b) fusion of selected Tokens$_t$ with Tokens$_{t-1}$ into Fused Tokens$_t$, where important patches use current frame tokens and others use previous frame tokens.}
\label{fig:patch_selection}
\end{figure}

Our fusion framework integrates two complementary analytical dimensions to identify important patches, as shown in Figure~\ref{fig:patch_selection}(a). This dual-constraint approach ensures comprehensive coverage of both low-level visual dynamics and high-level semantic relevance.

\subsubsection{Grayscale Pixel Difference Detection}

The pixel-level dimension captures fine-grained spatial changes through efficient grayscale-based analysis~\cite{fasola2006real}, contributing to the identification of important patches as part of the dual-dimension detection in Figure~\ref{fig:patch_selection}(a). Our approach offers several advantages over token-space similarity metrics commonly used in caching systems.

We convert RGB frames to grayscale using standard luminance weights to focus on brightness changes while maintaining computational simplicity:
\begin{equation}
\mathbf{G}_t = 0.299 \cdot \mathbf{I}_t^R + 0.587 \cdot \mathbf{I}_t^G + 0.114 \cdot \mathbf{I}_t^B
\end{equation}

This conversion is motivated by robotic manipulation scenarios where meaningful changes (object movements, shadow variations, lighting shifts) are primarily reflected in luminance rather than chromatic information.\cite{yang2012shadow, smeulders2014visual, secci2023rgb}

For each patch $i$ spanning spatial region $(u_i, v_i)$ to $(u_i + 13, v_i + 13)$ in the $14\times 14$ pixel patches, we compute the average absolute difference:
\begin{equation}
d_i^{\text{pixel}} = \frac{1}{196} \sum_{(u,v) \in \text{patch}_i} |\mathbf{G}_t(u,v) - \mathbf{G}_{t-1}(u,v)|
\end{equation}

Compared to cosine similarity on high-dimensional patch tokens, our pixel-based approach offers: (1) $\mathcal{O}(1)$ complexity per patch vs. $\mathcal{O}(d)$ for $d$-dimensional tokens, (2) direct interpretability of change magnitude, and (3) sensitivity to subtle manipulator movements that may be lost in token compression.

The grayscale pixel difference mask $m_i^{\text{pixel}} \in \{0,1\}$ indicates whether patch $i$ exhibits significant spatial changes, where 1 denotes using current frame tokens. The binary mask is computed as: $m_i^{\text{pixel}} = \mathbb{I}\!\left[d_i^{\text{pixel}} > \tau_{\text{pixel}}\right]$, where the pixel threshold $\tau_{\text{pixel}}$ is set based on scene statistics.

\subsubsection{Attention-Based Semantic Relevance Detection}

The attention dimension identifies semantically important patches using transformer attention patterns. Building upon token selection methods in vision-language models~\cite{chen2024image,zhang2024sparsevlm} and attention-guided approaches in vision transformers~\cite{dynamicvit2021,adavit2022,evit2022,tome2023,xu2025vla}, our approach introduces dual attention sources specifically adapted for temporal fusion in robotic manipulation tasks.

\begin{algorithm}
\caption{Temporal Token Fusion}
\begin{algorithmic}
\STATE \textbf{Input:} Current frame $\mathbf{I}_t$, previous frame $\mathbf{I}_{t-1}$, previous tokens $\mathbf{T}_{t-1}$, previous attention $\mathbf{A}_{t-1}$
\STATE \textbf{Output:} Fused tokens $\tilde{\mathbf{T}}_t$
\IF{IsKeyframe($t$)}
    \STATE $\tilde{\mathbf{T}}_t \leftarrow \text{VisionEncoder}(\mathbf{I}_t)$
\ELSE
    \STATE $\mathbf{T}_t, \mathbf{A}_t \leftarrow \text{VisionEncoder}(\mathbf{I}_t)$ \COMMENT{Extract tokens and attention weights}
    \FOR{$i = 1$ to $N$}
        \STATE $m_i^{\text{pixel}} \leftarrow \text{GrayscalePixelDiff}(\mathbf{I}_t, \mathbf{I}_{t-1}, i)$
        \STATE $m_i^{\text{attention}} \leftarrow \text{AttentionRelevance}(\mathbf{A}_{t-1}, i)$
        \STATE $m_i^{\text{fusion}} \leftarrow m_i^{\text{pixel}} \lor m_i^{\text{attention}}$
        \STATE $\tilde{\mathbf{t}}_t^{(i)} \leftarrow \text{TemporalFuse}(\mathbf{t}_t^{(i)}, \mathbf{t}_{t-1}^{(i)}, m_i^{\text{fusion}})$
    \ENDFOR
\ENDIF
\end{algorithmic}
\end{algorithm}

We extract attention weights $\mathbf{A}_{t-1}^{(l)}$ from a selected transformer layer and compute patch relevance through two complementary attention sources:

\textbf{Text-to-Vision Attention:} Captures semantic relevance based on task instruction by aggregating attention weights from text tokens to vision patches:
\begin{equation}
\mathbf{S}_{\text{text}}^{(l)} = \frac{1}{N_h} \sum_{h=1}^{N_h} \frac{1}{N_{\text{text}}} \sum_{j \in \text{text tokens}} \big(\mathbf{A}_{t-1}^{(l)}\big)_{h,\, j,\, \text{vision}}
\end{equation}

\textbf{Action-to-Vision Attention:} Leverages attention from the first action token, which encodes high-level manipulation strategy and spatial relevance for end-effector positioning:
\begin{equation}
\mathbf{S}_{\text{action}}^{(l)} = \frac{1}{N_h} \sum_{h=1}^{N_h} \big(\mathbf{A}_{t-1}^{(l)}\big)_{h,\, \text{action}_1,\, \text{vision}}
\end{equation}

For the selected layer $l$, we obtain the final task-relevance attention scores by selecting either text-to-vision or action-to-vision mode:
\begin{equation}
S_i^{\text{task}} = \mathbf{S}_{\text{mode}}^{(l)}[i]
\end{equation}

To avoid computational overhead during inference, we utilize attention weights from the previous timestep $\mathbf{A}_{t-1}$, motivated by the temporal stability of task-relevant regions in robotic manipulation. The attention mask $m_i^{\text{attention}} \in \{0,1\}$ is determined via top-k selection: $m_i^{\text{attention}} = \mathbb{I}[i \in \text{TopK}(S^{\text{task}}, k)]$, where patches with high attention scores (mask=1) use current frame tokens while others (mask=0) reuse historical tokens.

\section{Experiments}

\subsection{Experimental Setup}

\subsubsection{Baseline Models} We evaluate our temporal token fusion approach across multiple VLA architectures to demonstrate its model-agnostic applicability:
\begin{itemize}
\item \textbf{OpenVLA:} For LIBERO experiments, we employ officially released task-specific fine-tuned checkpoints (openvla-7b-finetuned-libero-\{object, spatial, goal, long\}), each optimized for the corresponding task suite. For SimplerEnv, we use the base OpenVLA-7B model without task-specific fine-tuning to evaluate cross-environment generalization. For real robot experiments, we fine-tune OpenVLA-7B on task-specific demonstration data.
\item \textbf{VLA-Cache:} We incorporate this recent architecture to demonstrate TTF's cross-model generalizability across different VLA paradigms. Notably, VLA-Cache's existing Key-Value matrix reuse mechanism provides an ideal testbed for exploring TTF's impact on Query matrix reuse, revealing promising directions for direct KQV matrix reuse strategies that could achieve computational acceleration while improving task performance.
\end{itemize}

\begin{table*}[t]
\centering
\small
\begin{tabular}{l|cccc|c}
\hline
Base Model & Object & Spatial & Goal & Long & Average \\
\hline
OpenVLA & 66.5 & 82.0 & 77.0 & 48.0 & 68.4 \\
OpenVLA + TTF & \textbf{72.5} & \textbf{84.5} & \textbf{79.0} & \textbf{53.5} & \textbf{72.4} \\
Fusion Rates & 44.6 & 41.2 & 43.5 & 42.0 & 42.8 \\
Improvement & +6.0 (+9.0\%) & +2.5 (+3.0\%) & +2.0 (+2.6\%) & +5.5 (+11.5\%) & +4.0 (+5.8\%) \\
\hline
VLA-Cache & 69.0 & 84.0 & 77.0 & 55.0 & 71.3 \\
VLA-Cache + TTF & \textbf{73.0} & \textbf{84.0} & \textbf{81.0} & \textbf{58.0} & \textbf{74.0} \\
Improvement & +4.0 (+5.8\%) & +0.0 (+0.0\%) & +4.0 (+5.2\%) & +3.0 (+5.5\%) & +2.7 (+3.8\%) \\
\hline
\end{tabular}
\caption{Task success rates (\%) and temporal fusion rates (\%) on the LIBERO benchmark using OpenVLA models fine-tuned for each task suite. Success rates denote the percentage of successful task completions across 200 episodes per task suite (10 tasks × 20 episodes each). The fusion rates represent the proportion of vision tokens reused from the previous frame in the fused representation.}
\label{tab:libero_results}
\end{table*}

\subsubsection{Evaluation Benchmarks} Our comprehensive evaluation spans both simulation and real-world environments:
\begin{itemize}
\item \textbf{LIBERO:} Simulation evaluation across four task suites: (1) \textit{Object} - manipulation of diverse objects (e.g., varying shapes, sizes) through single-object interactions; (2) \textit{Spatial} - tasks requiring precise spatial reasoning and object placement relative to landmarks; (3) \textit{Goal} - complex goal-conditioned tasks with multi-step reasoning for specific configurations; (4) \textit{Long} - long-horizon tasks testing temporal consistency over sequential manipulations. Each suite contains 10 distinct tasks with 20 evaluation episodes per task (200 total episodes per suite).

\item \textbf{SimplerEnv:} Simulation benchmark for evaluating real-world robot manipulation policies across diverse scenarios. We evaluate three representative tasks: \textit{Move Near Object} (240 episodes), \textit{Pick Coke Can} with multiple orientations (300 episodes), and \textit{Drawer Operations} (216 episodes), providing comprehensive validation across varying manipulation complexities and interaction types.

\item \textbf{Real Robot Tasks:} Physical validation using a Franka Research 3 robot across three manipulation tasks spanning different complexities: single-object pick-and-place (\textit{put the garlic on the plate}), multi-object sequential manipulation (\textit{put the pepper and corn on the plate}), and contact-rich manipulation (\textit{close the drawer}). We collect 80 demonstration episodes per task using Gello teleoperation at 5 Hz, and fine-tune OpenVLA-7B for 20,000 steps with a batch size of 8. During evaluation, the fine-tuned models are deployed at 5 Hz and tested with 20 episodes for each task.

\end{itemize}

\subsection{Main Results}

\subsubsection{LIBERO Experiments}

\begin{figure*}[h]
\centering
\includegraphics[width=\textwidth]{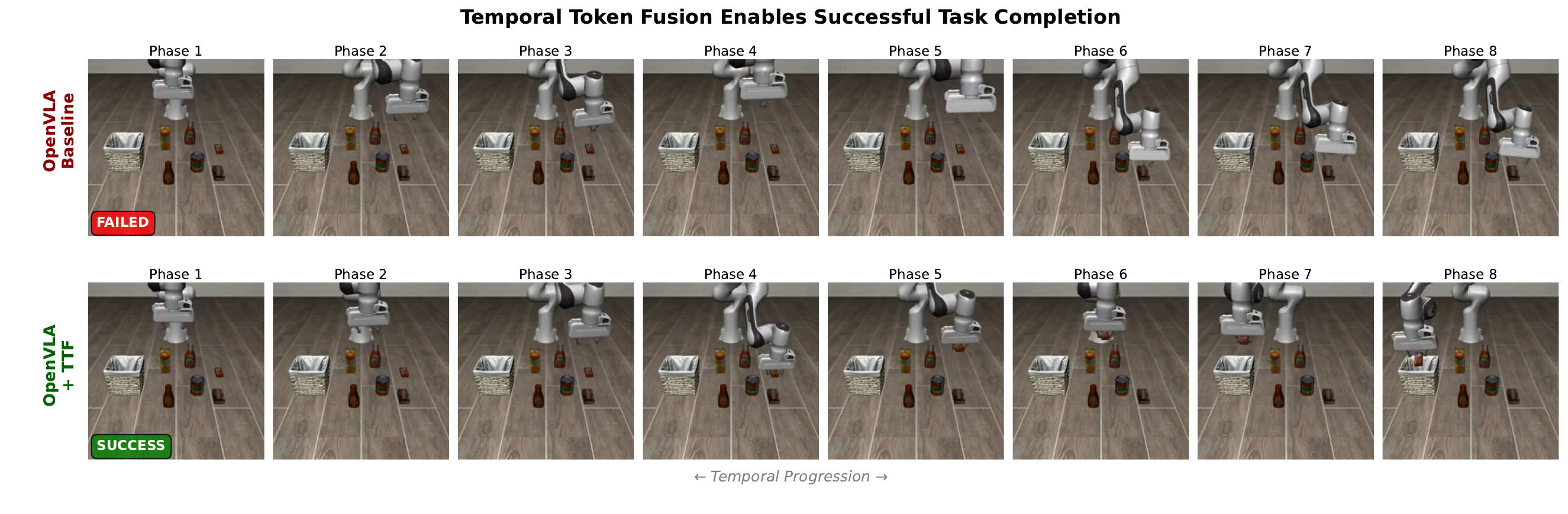}
\caption{Temporal progression analysis illustrating a failure-to-success transition for the task instruction ``pick up the butter and place it in the basket.'' Eight key phases showing OpenVLA baseline (failed) vs. OpenVLA + TTF (successful), demonstrating the critical role of temporal consistency in successful manipulation.}
\label{fig:baseline_tff_comparison}
\end{figure*}

Table~\ref{tab:libero_results} presents our core experimental results across four task suites using two representative VLA architectures, demonstrating the effectiveness and model-agnostic applicability of our temporal token fusion approach. Our analysis reveals several key insights: \textit{Model-agnostic effectiveness}: Both OpenVLA and VLA-Cache demonstrate consistent improvements (4.0 and 2.7 percentage points average respectively), validating the generalizability of our approach across different VLA architectures. \textit{Task-specific patterns}: Long-horizon tasks benefit most significantly from temporal fusion (+11.5\% relative improvement for OpenVLA), suggesting that extended manipulation sequences particularly benefit from temporal context integration, while object manipulation shows strong absolute gains (+6.0 percentage points for OpenVLA). \textit{Fusion efficiency}: The fusion rates (42.8\% average for OpenVLA) indicate substantial feature reuse while maintaining performance gains, demonstrating that our dual-dimension detection successfully identifies stable regions for temporal integration without compromising inference quality. \textit{Temporal progression analysis}: Figure~\ref{fig:baseline_tff_comparison} illustrates representative failure-to-success cases where baseline methods fail but TTF succeeds, highlighting the critical role of temporal coherence in successful robotic manipulation task completion. TTF introduces less than 2\% additional runtime overhead, confirming its efficiency and suitability for real-time robotic inference.

\textbf{Implicit Query Reuse and Noise Robustness:} A particularly revealing insight emerges from VLA-Cache+TTF results. VLA-Cache~\cite{xu2025vla} accelerates inference by reusing Key-Value matrices ($\mathbf{K}_{t-1}^{(l)}, \mathbf{V}_{t-1}^{(l)}$) for static visual patches while always recomputing Query matrices ($\mathbf{Q}_t^{(l)}$) to maintain contextual sensitivity. However, when TTF is applied to VLA-Cache, an important mechanism emerges: TTF's token-level fusion ($\tilde{\mathbf{t}}_t^{(i)} = \mathbf{t}_{t-1}^{(i)}$ for selected patches) means that the corresponding Query matrix portions are approximately and implicitly reused since $\mathbf{Q}_t^{(l)} = \mathbf{W}_q^{(l)} \cdot \tilde{\mathbf{T}}_t$. Thus, VLA-Cache+TTF effectively \textit{nearly reuses all three attention matrices} (K, V, and Q) for static patches, going beyond VLA-Cache's original KV-only reuse strategy. Contrary to conventional wisdom that Query reuse degrades performance due to contextual sensitivity, our results show significant improvements, particularly for long-horizon tasks (+3.0 points: 55.0\%→58.0\%). This performance improvement demonstrates that our dual-dimension detection provides stabilized contextual representations that enhance robustness against visual observation noise—including lighting fluctuations, motion blur, and sensor artifacts common in robotic manipulation. The selective Query reuse validates TTF's core hypothesis: temporal coherence in stable regions enhances rather than compromises inference quality. This finding reveals a promising ``free lunch'' future direction worth exploring: \textit{direct KQV matrix reuse} for static patches could achieve computational acceleration while simultaneously improving task success rates.

\textbf{Configuration Strategy:} Parameter selection demonstrates TTF's robustness. OpenVLA employs consistent parameters: keyframe interval K=3, attention top-k=70, pixel threshold=0.03, and text-to-vision attention. VLA-Cache requires minimal adaptation with task-adaptive fusion rates (30\% for Object/Spatial/Long, 50\% for Goal tasks). Since VLA-Cache uses patch importance scoring, we directly utilize its scores for patch selection, avoiding redundant computation. Neither model requires task-specific hyperparameter tuning.

\subsubsection{SimplerEnv Cross-Environment Validation}

\begin{table}[h]
\centering
\small
\begin{tabular}{l|cc|cc}
\hline
Task & \multicolumn{2}{c|}{Success Rate (\%)} & \multicolumn{2}{c}{Improvement} \\
& OpenVLA & TTF & Absolute & Relative \\
\hline
Move Near & 49.4 & 52.1 & +2.7 & +5.5\% \\
Pick Coke & 17.0 & 18.7 & +1.7 & +10.0\% \\
Drawer & 33.3 & 33.8 & +0.5 & +1.5\% \\
\hline
\textit{Average} & 33.2 & 34.9 & +1.6 & +4.8\% \\
\hline
\end{tabular}
\caption{Cross-environment validation on SimplerEnv benchmark using the base OpenVLA-7B model (without task-specific fine-tuning) with identical TTF parameters as LIBERO experiments. Results demonstrate consistent generalization across simulation platforms and manipulation complexities.}
\label{tab:simplerenv_results}
\end{table}

To validate cross-environment generalization, we evaluate TTF on SimplerEnv using identical parameters as our LIBERO experiments. The experimental results in Table~\ref{tab:simplerenv_results} demonstrate consistent improvements across all three tasks, with an average improvement of 1.6 percentage points (4.8\% relative). TTF shows particularly strong gains in Pick Coke Can (+10.0\% relative), indicating that complex grasping tasks with multiple orientations benefit from temporal coherence.

These results validate TTF's environment-agnostic effectiveness, confirming that temporal coherence benefits extend across different simulation platforms without requiring parameter re-tuning.

\begin{table*}[t]
\centering
\small
\begin{tabular}{m{2.6cm}<{\centering}|cccc|c|c}
\hline
Method & Object & Spatial & Goal & Long & Average & Fusion Rates \\
\hline
OpenVLA & 66.5 & 82.0 & 77.0 & 48.0 & 68.4 & -- \\
\multirow{2}{*}{Pixel-only TTF} & 72.0 & 80.5 & 76.5 & 52.5 & 70.4 & \multirow{2}{*}{61.1/57.8/59.9/59.8} \\
& +5.5 (+8.3\%) & -1.5 (-1.8\%) & -0.5 (-0.6\%) & +4.5 (+9.4\%) & +2.0 (+2.9\%) & \\
\multirow{2}{*}{Attention-only TTF} & 68.0 & 83.0 & 77.5 & 56.5 & 71.3 & \multirow{2}{*}{48.3/48.2/48.3/48.4} \\
& +1.5 (+2.3\%) & +1.0 (+1.2\%) & +0.5 (+0.6\%) & +8.5 (+17.7\%) & +2.9 (+4.2\%) & \\
\multirow{2}{*}{\textbf{Pixel-Attention TTF}} & \textbf{72.5} & \textbf{84.5} & \textbf{79.0} & \textbf{53.5} & \textbf{72.4} & \multirow{2}{*}{\textbf{44.6/41.2/43.5/42.0}} \\
& +6.0 (+9.0\%) & +2.5 (+3.0\%) & +2.0 (+2.6\%) & +5.5 (+11.5\%) & +4.0 (+5.8\%) & \\
\hline
\end{tabular}
\caption{Task success rates (\%) and fusion rates (\%) in analysis dimension ablation study across LIBERO task suites using OpenVLA task-specific fine-tuned checkpoints. Success rates denote the percentage of successful completions across 200 episodes per task suite (10 tasks × 20 episodes each). Fusion rates represent the proportion of vision tokens from the previous frame retained in the fused representation.}
\label{tab:dimension_ablation}
\end{table*}

\subsubsection{Real Robot Experiments}

\begin{figure}[h]
\centering
\includegraphics[width=\columnwidth]{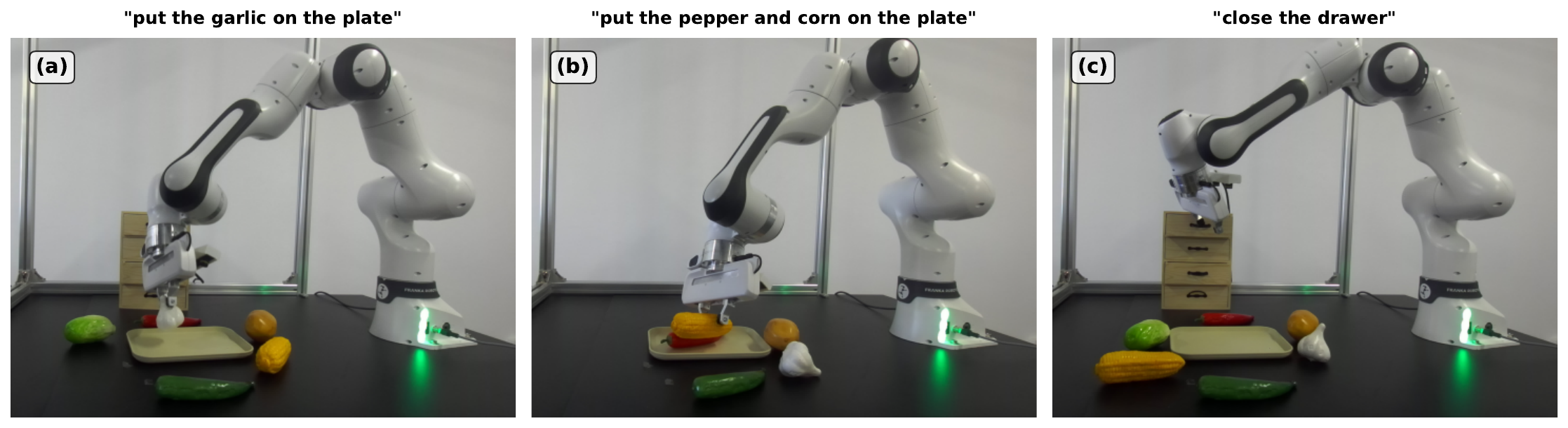}
\caption{Real robot manipulation tasks used for physical validation of TTF: (a) single-object pick-and-place, (b) multi-object sequential manipulation, and (c) contact-rich drawer closing.}
\label{fig:real_robot_tasks}
\end{figure}

\begin{table}[h]
\centering
\small
\begin{tabular}{l|cc|cc}
\hline
Task & \multicolumn{2}{c|}{Success Rate (\%)} & \multicolumn{2}{c}{Improvement} \\
& OpenVLA & TTF & Absolute & Relative \\
\hline
Garlic & 40.0 & 45.0 & +5.0 & +12.5\% \\
Pepper \& Corn & 30.0 & 35.0 & +5.0 & +16.7\% \\
Drawer & 45.0 & 45.0 & +0.0 & +0.0\% \\
\hline
\textit{Average} & 38.3 & 41.7 & +3.3 & +8.7\% \\
\hline
\end{tabular}
\caption{Real robot manipulation success rates (\%) across 20 episodes per task using OpenVLA-7B fine-tuned on task-specific data. Results demonstrate TTF's effectiveness in noisy real-world environments with enhanced temporal stability benefits.}
\label{tab:real_robot_results}
\end{table}

For real robot experiments, we employ more sensitive parameter settings to handle visual noise: pixel threshold=0.01 and attention top-k=100, while maintaining other configurations consistent with LIBERO experiments. Real-world environments provide an ideal testbed for TTF's noise robustness due to dynamic lighting, sensor noise, motion blur, and other visual artifacts absent in simulation.

The experimental results in Table~\ref{tab:real_robot_results} demonstrate TTF's effectiveness in real-world scenarios, achieving an average improvement of 3.3 percentage points (8.7\% relative). Both pick-and-place tasks (Garlic, Pepper \& Corn) show substantial gains from temporal coherence (+5.0 points each, 12.5\% and 16.7\% relative), demonstrating TTF's particular strength in visual tracking and object manipulation tasks. These results validate TTF's practical applicability in real-world deployment, especially for manipulation scenarios involving object interaction and spatial reasoning.

\subsection{Ablation Studies}

We conduct comprehensive ablation studies to validate the necessity of each component in our dual-dimension temporal fusion approach.

\subsubsection{Analysis Dimension Validation}

Table~\ref{tab:dimension_ablation} validates the necessity of our dual-dimension detection approach. Pixel-based analysis excels in detecting spatial changes, while attention-based analysis better captures task-relevant semantics. Our hybrid method delivers the best average performance (72.4\%) by adaptively integrating both dimensions. Notably, its conservative OR-based fusion logic results in the lowest fusion rates (42.8\%), prioritizing inference quality and confirming the complementary nature of pixel and attention-based detection.

\subsubsection{Keyframe Mechanism Analysis}

\begin{figure}[h]
\centering
\includegraphics[width=\columnwidth]{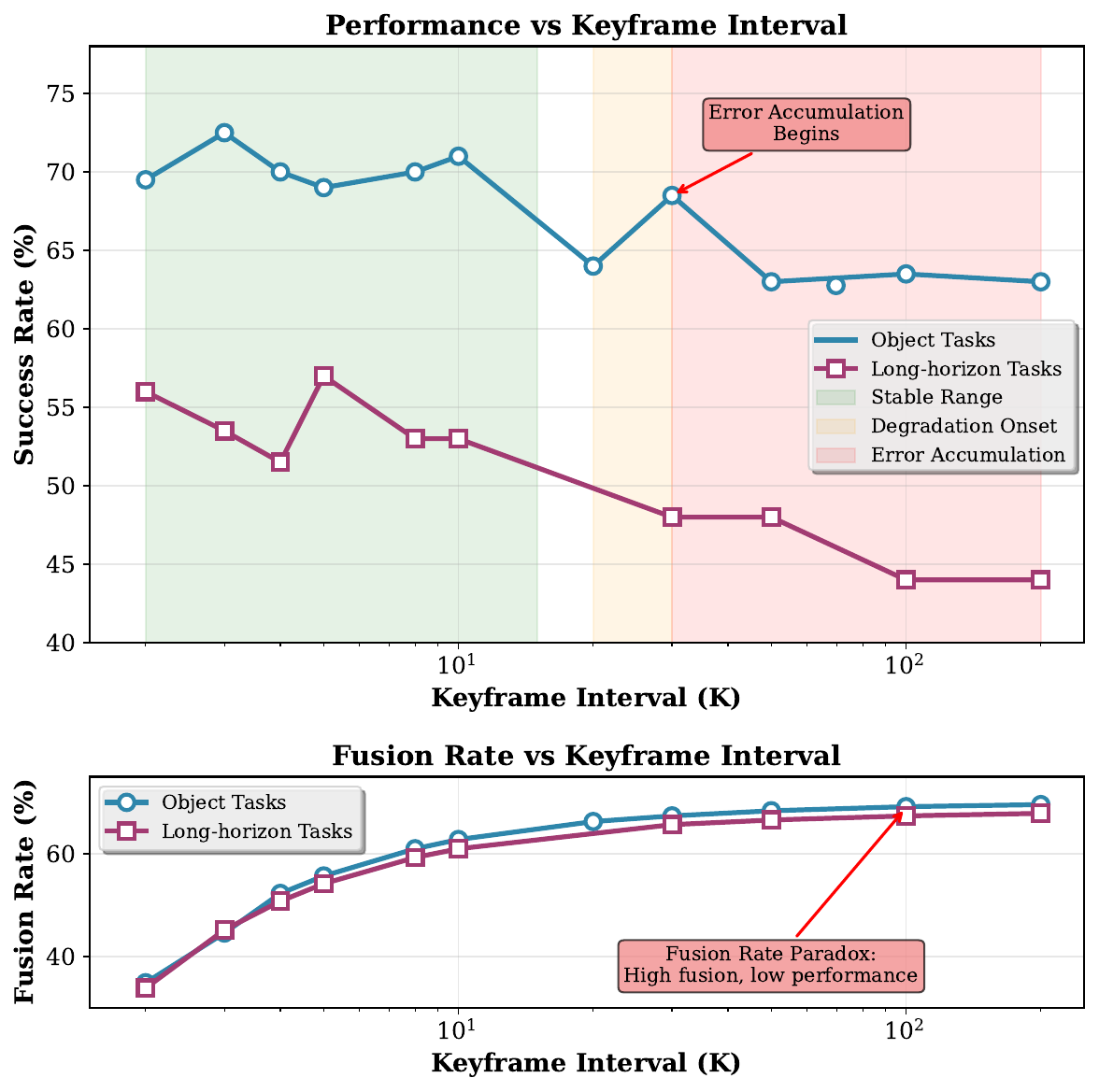}
\caption{Keyframe interval analysis across Object and Long task suites. Top: Performance vs. Keyframe Interval showing error accumulation beyond K=30. Bottom: Fusion Rates vs. Keyframe Interval revealing the efficiency-performance trade-off. The analysis demonstrates three distinct regimes: stable performance (K$\leq$15), degradation onset (K=20--30), and error accumulation (K$\geq$30).}
\label{fig:keyframe_analysis}
\end{figure}

Figure~\ref{fig:keyframe_analysis} presents keyframe interval analysis across 14 configurations (K=2 to K=200), revealing three distinct regimes: \textit{Stable Range} (K$\leq$15) with optimal performance at K=3, \textit{Degradation Onset} (K=20-30) where error accumulation begins, and \textit{Error Accumulation} (K$\geq$30) with performance plateau. Long-horizon tasks show higher sensitivity to temporal drift. Fusion rates increase monotonically with keyframe interval (from ~35\% to ~70\%), highlighting the efficiency-performance trade-off and validating our conservative dual-dimension approach.

\section{Conclusion}

We present Temporal Token Fusion (TTF), a training-free approach that addresses the fundamental limitation of current VLA models in leveraging temporal coherence during robotic manipulation. Our dual-dimension detection framework combines pixel-level spatial dynamics with attention-based semantic relevance assessment to intelligently integrate historical and current visual representations. Comprehensive evaluation demonstrates consistent improvements across LIBERO (4.0 percentage points average), SimplerEnv (cross-environment generalization), and real robot tasks (8.7\% relative improvement), with model-agnostic applicability across OpenVLA and VLA-Cache architectures. Beyond performance gains, our work reveals the unexpected benefits of selective Query matrix reuse in attention mechanisms, suggesting promising directions for direct KQV matrix strategies that achieve computational acceleration while improving task success rates. 

\bibliography{aaai2026}

\end{document}